\documentclass[letterpaper]{article} % DO NOT CHANGE THIS
\usepackage[preprint]{aaai2027} % Non-anonymous arXiv preprint; retain the AAAI layout.
\usepackage[hyphens]{url}  % DO NOT CHANGE THIS
\usepackage{graphicx} % DO NOT CHANGE THIS
\usepackage{natbib}  % DO NOT CHANGE THIS AND DO NOT ADD ANY OPTIONS TO IT
\usepackage{caption} % DO NOT CHANGE THIS AND DO NOT ADD ANY OPTIONS TO IT
\usepackage{amsmath,amsfonts,amssymb}
\usepackage{booktabs}
\usepackage{array}
\usepackage{multirow}
\usepackage{colortbl}
\usepackage{makecell}
\usepackage{placeins}

\definecolor{lightpink}{RGB}{255, 230, 230}
\definecolor{lightblue}{RGB}{220, 220, 220}
\definecolor{graytext}{gray}{0.6}
\newcolumntype{A}{>{\columncolor{lightblue}}c}
\newcommand{\conf}[1]{\textcolor{graytext}{\textit{\scriptsize{#1}}}}

\title{PRISM: Primitive-Integrated Separable Modulation with Priority Visual Memory Queue for Compositional Zero-Shot Learning}
\author{
    Jianhua Wang\textsuperscript{\rm 1}\equalcontrib,
    Zhihang Zheng\textsuperscript{\rm 2}\equalcontrib,
    Yang Liu\textsuperscript{\rm 3}\corresponding
    Jiale Du\textsuperscript{\rm 3}
}
\affiliations{
    \textsuperscript{\rm 1}School of Electronic Engineering, Xidian University, Shaanxi 710071, China\\
    \textsuperscript{\rm 2}School of Computer Science and Technology, Xidian University, Shaanxi 710071, China\\
    \textsuperscript{\rm 3}School of Telecommunications Engineering, Xidian University, Shaanxi 710071, China\\
    \textsuperscript{\rm 3}yliu1@xidian.edu.cn
}

\begin{document}

\maketitle

\begin{abstract}
Compositional Zero-Shot Learning (CZSL) recognizes unseen attribute--object pairs by recombining primitives from seen compositions. Existing methods stop sharing evidence at deployment: fixed prototypes cannot exploit unlabeled test streams, while composition-wise refinement cannot propagate corrections across pairs sharing a primitive. We introduce PRISM, an online CZSL framework that extends primitive sharing to test-time adaptation. Its Hierarchical Knowledge Accumulation Module (HiKAM) factorizes prototype migration into composition-, attribute-, and object-level residual banks. A Priority Visual Memory Queue (PVMQ), initialized by Prototype Queue Initialization (PQI), grounds text updates with bounded visual evidence. The same adaptive core supports exhaustive Full PRISM and recall-calibrated Adaptive PRISM-Lite, which selects candidate budgets from held-out ranks. Comprehensive closed- and open-world experiments on UT-Zappos, MIT-States, and C-GQA demonstrate the state of the art, with Full PRISM achieving the best open-world AUC and harmonic mean on all three benchmarks and the best closed-world results on UT-Zappos and C-GQA. Fine-grained long-tail and non-stationary primitive-block evaluations reveal gains on under-supported concepts and stable adaptation under prolonged skew. Adaptive PRISM-Lite further delivers substantial efficiency gains in the full open-world space with only a small accuracy cost. PRISM thus combines primitive-shared online adaptation, multimodal grounding, and candidate-aware inference.
\end{abstract}

%\begin{links}
%    \link{Code}{https://aaai.org/example/code}
%    \link{Datasets}{https://aaai.org/example/datasets}
%    \link{Extended version}{https://aaai.org/example/extended-version}
%\end{links}

\section{Introduction}
\label{sec:introduction}
Humans recognize unseen compositions by recombining familiar attributes and objects. Compositional Zero-Shot Learning (CZSL) formalizes this ability by learning from seen attribute--object pairs and recognizing unseen recombinations~\cite{atzmon2016learning,misra2017redwine}. Generalized and open-world protocols extend this task from a restricted set of evaluation pairs to much larger Cartesian-product candidate spaces containing many implausible compositions~\cite{AUC_purushwalkam2019taskdrivenmodularnetworkszeroshot,Co-CGE}. This expansion is particularly difficult when compositional data are long-tailed: frequent pairs dominate the available evidence, while rare compositions and under-supported attributes or objects remain difficult to distinguish. Primitive sharing is therefore central to CZSL because evidence about one attribute or object can support every composition containing that primitive, rather than treating each pair as an independent category.

\begin{figure}[!t]
    \centering
    \includegraphics[width=\linewidth]{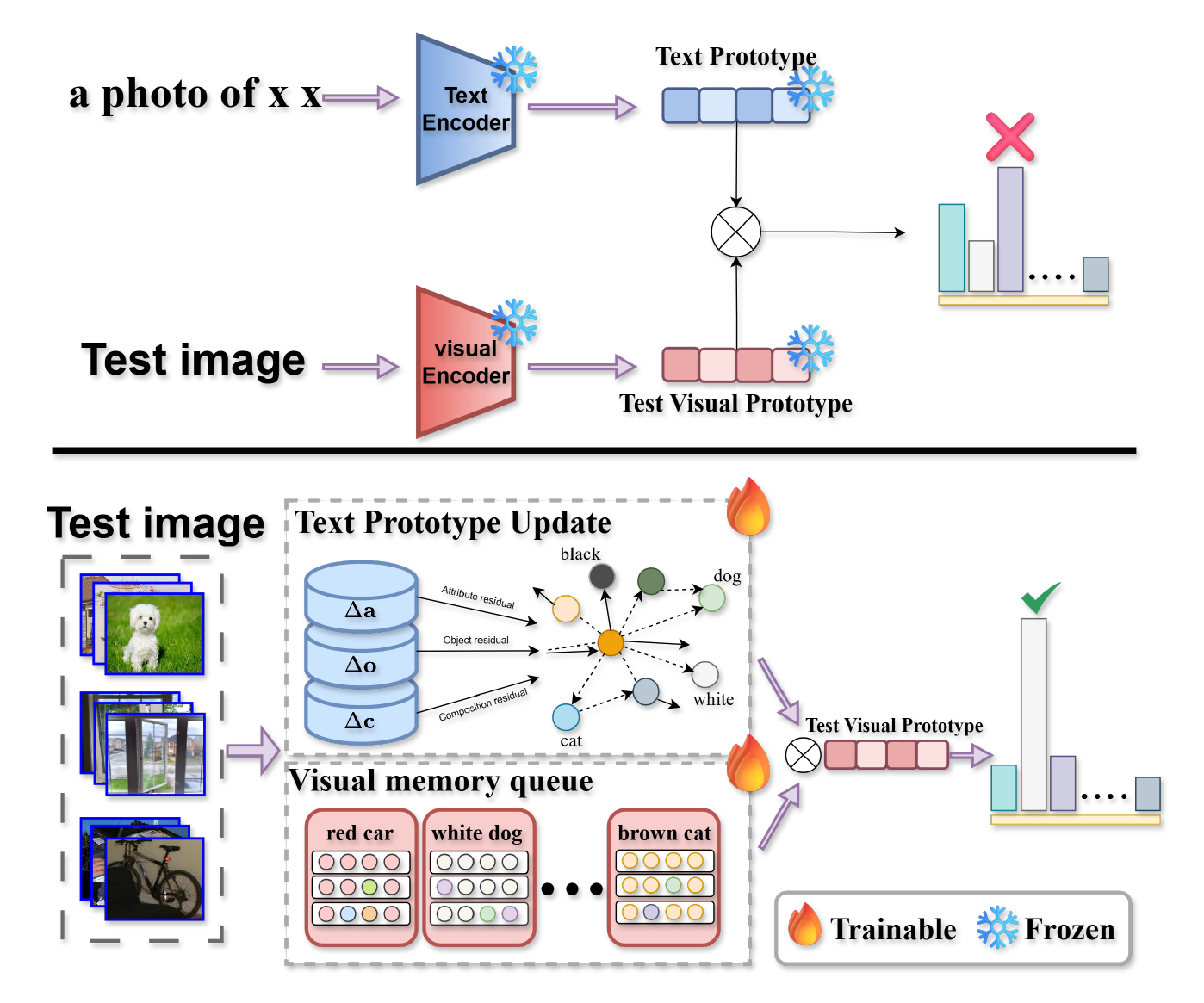}
    \caption{Test-time adaptation under label-space shift, showing primitive-shared textual updates and Priority Visual Memory Queue retrieval from an unlabeled test stream.}
    \label{fig:prism_motivation}
\end{figure}

Existing CZSL methods mainly realize primitive sharing during offline representation learning through structured factorization, graph propagation, and vision--language prompt or prototype learning~\cite{nagarajan2018attributes,GIPCOL,ClusPro}. Figure~\ref{fig:prism_motivation} contrasts fixed deployment with primitive-shared test-time adaptation. In an unlabeled test stream, incoming images expose visual regularities unavailable in seen training data, but static prototypes cannot incorporate this evidence, and composition-wise refinement cannot transfer a correction to pairs sharing the same attribute or object. These limitations become most consequential in two demanding regimes: open-world recognition, where candidate-dependent confusion and computation grow rapidly, and fine-grained long-tail recognition, where rare pairs provide too little direct evidence for independent adaptation. Text-only self-training~\cite{liang2020source} may further amplify noisy pseudo-labels. Effective online CZSL therefore requires an update rule that preserves primitive sharing, grounds migration with visual evidence, and controls candidate-dependent computation. We term this setting \textbf{Compositional Test-Time Adaptation}: online adaptation that explicitly preserves compositional primitive sharing, which generic TTA objectives do not directly encode. Additionally, efficiency is essential in practical scenarios, where sequential adaptation must remain time-efficient as recurring prototype migration and memory matching scale with the candidate-space size.

To address this gap, we propose \textbf{PRISM} (\textbf{Pr}imitive-\textbf{I}ntegrated \textbf{S}eparable \textbf{M}odulation with Priority Visual Memory Queue), extending attribute and object sharing from offline representation learning into online test-time adaptation. Its Hierarchical Knowledge Accumulation Module (HiKAM) factorizes prototype migration into composition-, attribute-, and object-level residual banks, letting evidence from one stream sample propagate to related pairs while preserving pair-specific corrections. The Priority Visual Memory Queue (PVMQ) grounds this migration in bounded, class-indexed visual evidence, with Prototype Queue Initialization (PQI) supplying visual anchors ahead of reliable unseen pseudo-labels. Only lightweight residual variables are updated, keeping the CLIP~\cite{clip} encoders and learned prompt interface fixed. This adaptive core supports two complementary routes---Full PRISM, which exhaustively processes the complete candidate space, and Adaptive PRISM-Lite, which applies validation-rank-calibrated selection to cut recurring migration and matching cost---making the accuracy--efficiency trade-off explicit. Evaluation accordingly separates task-level positioning via published CZSL comparisons from adaptation-specific evidence isolated through matched No-TTA and component ablations.

The main contributions of this work are as follows:
\begin{itemize}
    \item \textbf{Primitive-shared, visually grounded online adaptation.} HiKAM factorizes prototype migration into composition-, attribute-, and object-level residual banks, extending primitive sharing from offline learning to the online update rule with the CLIP encoders kept frozen. PVMQ and PQI ground this migration with bounded, class-indexed visual evidence, letting stream evidence support related attribute--object pairs without relying on text-only self-training.
    
    \item \textbf{Candidate-controlled efficient inference.} Adaptive PRISM-Lite applies validation-rank calibration to prune candidate-dependent computation, delivering large throughput and latency gains in the open-world setting at a marginal cost to accuracy relative to Full PRISM.
    
    \item \textbf{State-of-the-art empirical performance.} Full PRISM achieves the highest reported AUC and HM on all three benchmarks under the most demanding open-world setting, along with the strongest closed-world results on UT-Zappos and C-GQA, with matched same-checkpoint ablations confirming positive contributions from the online path.
    
    \item \textbf{Fine-grained long-tail evaluation.} We introduce an evaluation protocol that disaggregates accuracy by composition support and primitive frequency, revealing substantial gains on under-supported compositions and primitives.
\end{itemize}
\section{Related Work}
\label{sec:related_work}

\paragraph{Compositional Representation Learning.}
CZSL methods learn reusable attribute and object primitives from seen pairs and recombine them for unseen-pair recognition. Early approaches either construct pair representations with contextual composition operators or decompose images into attribute- and object-oriented branches~\cite{atzmon2016learning,misra2017redwine,nagarajan2018attributes}. Structured variants impose causal objectives, symmetry, graph propagation, or conditional dependence to regulate how the two primitives interact~\cite{atzmon2020causal,li2020symmetry,Wu2025CPF}. With pretrained vision--language models, trainable or decomposed prompts, graph-injected prompts, and multi-path fusion align images with composition- and primitive-level text embeddings~\cite{CSP,GIPCOL,Troika}. These mechanisms share primitive knowledge while learning the offline representation, but the resulting interface is generally fixed during deployment. PRISM preserves the offline recognizer and extends the same sharing principle to the online update rule through composition-, attribute-, and object-level residual banks.

\paragraph{Primitive and Multimodal Prototype Learning.}
Prototype-based CZSL methods refine the semantic anchors used to compare images with candidate pairs. Distributional priors and retrieval augment a prototype with linguistic context, while multi-path or complex-valued representations preserve complementary composition and primitive evidence~\cite{PLID,PAPR,Troika}. Clustering-based prototypes model multiple visual modes for a primitive rather than collapsing them into one center, and visual proxies reduce the image--text modality gap with trainable visual centers~\cite{ClusPro,Zhang2025VisualProxy}. These approaches improve prototype coverage and discrimination using training data, but their visual evidence remains fixed when an unlabeled test stream arrives. PRISM instead maintains bounded online visual state: PVMQ retains entropy-prioritized, class-indexed features, PQI supplies initial anchors, and multimodal alignment couples this evidence to the text prototypes migrated by HiKAM.

\paragraph{Online Test-Time Adaptation.}
Online test-time adaptation updates a pretrained model from unlabeled test samples through entropy minimization~\cite{tent,chen2022tent}, self-supervised objectives~\cite{sun2020testtime}, teacher-student distillation~\cite{wang2022continual}, augmentation and reliability filtering~\cite{zhang2022memo,niu2022eata,niu2023sar}, or prompt and reconstruction tuning~\cite{gandelsman2022ttmae,AUC_shu2022testtimeprompttuningzeroshot}. Applying these methods to CZSL introduces task-specific design choices. The standard Tent formulation updates normalization parameters commonly associated with Batch Normalization, whereas CLIP's ViT-L/14 uses Layer Normalization~\cite{tent,clip}. A CLIP-compatible implementation therefore requires an explicit choice of adapted variables. MEMO evaluates multiple augmentations per test sample~\cite{zhang2022memo}, increasing encoding and scoring cost in large candidate spaces. Their application to CZSL also requires a rule for propagating composition evidence across candidates sharing an attribute or object. PRISM encodes primitive sharing through composition-, attribute-, and object-level residual banks with bounded visual memory. Published CZSL results provide task-level context, and matched internal controls quantify PRISM's online contribution under a shared checkpoint and stream.
\begin{figure*}[t]
    \centering
    \includegraphics[width=\textwidth]{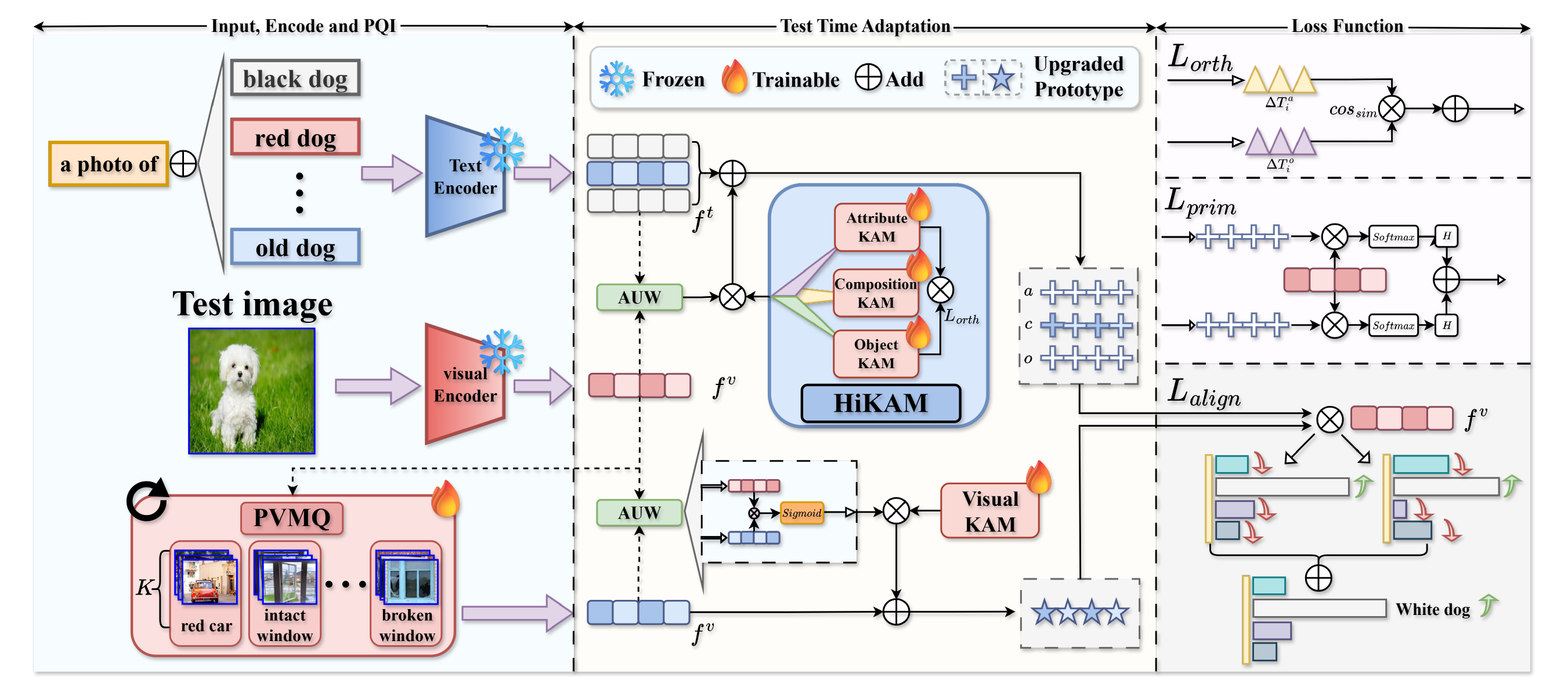}
    \caption{Overall architecture of PRISM at test time, showing primitive-shared textual and visual updates to composition prototypes.}
    \label{fig:PRISM_framework}
\end{figure*}

\section{Methodology}
\label{sec:method}

\subsection{Task Formulation}
\label{sec:task_formulation}

CZSL learns from seen attribute--object pairs and recognizes unseen recombinations. Let $\mathcal{A}=\{a_1,\ldots,a_M\}$ and $\mathcal{O}=\{o_1,\ldots,o_N\}$ denote the primitive vocabularies, with composition $c=(a,o)$. The training set $\mathcal{D}_{\mathrm{tr}}=\{(x_i,c_i)\}_{i=1}^{n}$ contains $\mathcal{C}^{s}$, while held-out compositions $\mathcal{C}^{u}$ satisfy $\mathcal{C}^{s}\cap\mathcal{C}^{u}=\emptyset$. We define the evaluation universe as $\mathcal{C}_{\mathrm{eval}}=\mathcal{C}^{s}\cup\mathcal{C}^{u}$ in the closed world and $\mathcal{C}_{\mathrm{eval}}=\mathcal{A}\times\mathcal{O}$ in the open world, where infeasible pairs may occur.

Figure~\ref{fig:PRISM_framework} summarizes PRISM's two-route test-time architecture: a recognizer trained on seen pairs freezes its CLIP encoders and learned prompts, updating only primitive-structured text residuals and bounded PVMQ evidence.

\subsection{Validation-Rank-Calibrated Dual-Route Inference}
\label{sec:dual_route_inference}

Full PRISM and Adaptive PRISM-Lite share the checkpoint, modules, objective, and online updates; only their active candidate sets differ. \textbf{Full PRISM} uses $\mathcal{C}_i=\mathcal{C}_{\mathrm{eval}}$, whereas \textbf{Adaptive PRISM-Lite} uses $\mathcal{C}_i=\mathcal{S}_i^{\widehat{N}}$ with budget $\widehat{N}$ calibrated once from held-out validation ranks. Candidate selection controls recurring computation and may alter subsequent adaptation trajectories.

For fixed base prototype $t_c^{0}$ and normalized image feature $\bar{v}_i$, Adaptive PRISM-Lite scores $q_{i,c}=\bar{v}_i^{\top}\bar{t}_c^{0}$ and retains
\begin{equation}
    \mathcal{S}_i^{\widehat{N}}
    =\left\{c\in\mathcal{C}_{\mathrm{eval}}:q_{i,c}\geq q_{i,(\widehat{N})}\right\},
    \label{eq:prism_lite_retrieval}
\end{equation}
where $q_{i,(\widehat{N})}$ is the $\widehat{N}$-th largest score. Boundary ties are retained, so $|\mathcal{S}_i^{\widehat{N}}|\geq\widehat{N}$.

Calibration uses no test labels. Let $r_j$ be the one-based ground-truth rank for validation sample $j$, with ordered ranks $r_{(1)}\leq\cdots\leq r_{(m)}$. For target recall $\tau_{\mathrm{rec}}\in(0,1)$, define $k=\lceil(m+1)\tau_{\mathrm{rec}}\rceil$ and
\begin{equation}
    \small
    \widehat{N}=
    \begin{cases}
        \min\!\left(|\mathcal{C}_{\mathrm{eval}}|,
        \max\!\left(N_{\min},r_{(k)}\right)\right), & k\leq m,\\
        |\mathcal{C}_{\mathrm{eval}}|, & k=m+1.
    \end{cases}
    \label{eq:prism_lite_calibration}
\end{equation}
Thus, $\widehat{N}$ is the finite-sample rank quantile clipped to $[N_{\min},|\mathcal{C}_{\mathrm{eval}}|]$, adapting the budget rather than transferring a fixed Top-$N$ across settings.

\paragraph{Conditional marginal coverage.}
Under exchangeable validation and test coarse ranks, the split-conformal order-statistic argument~\cite{angelopoulos2023conformal} gives, for test target $c_i^{\star}$,
\begin{equation}
    \Pr\!\left(c_i^{\star}\in\mathcal{S}_i^{\widehat{N}}\right)
    \geq \frac{k}{m+1}\geq\tau_{\mathrm{rec}}.
    \label{eq:prism_lite_coverage}
\end{equation}
Neither $N_{\min}$ nor boundary ties can reduce coverage. This bound guarantees coarse-retrieval inclusion under exchangeability, not final accuracy or AUC. Adaptive PRISM-Lite then applies AUW, HiKAM, and PVMQ only to $c\in\mathcal{S}_i^{\widehat{N}}$; attribute and object prototypes remain unpruned for primitive-confidence estimation. The fixed coarse scoring spans $\mathcal{C}_{\mathrm{eval}}$, while adaptive migration and matching scale with $\widehat{N}$.

\subsection{Training Phase: Base Composition Recognizer}
\label{sec:training_phase}

The base recognizer adapts CLIP to seen pairs using lightweight visual adapters and soft composition prompts~\cite{houlsby2019adapters,chen2022adaptformer,CSP}, with both pretrained transformers frozen. For $c=(a,o)$, shared context and learned primitive embeddings produce $t_c$, scored as $\gamma\bar v^{\top}\bar t_c$~\cite{CoOp}. Cross-entropy over $\mathcal{C}^{s}$ trains only the adapters and prompts, which are frozen before PRISM updates the test-time residual state. Full training details appear in the Technical Supplement.

\subsection{Test Phase: PRISM Online Adaptation}
\label{sec:test_phase}

Before testing, the frozen text encoder forms $T=[t_1,\ldots,t_{|\mathcal{C}_{\mathrm{eval}}|}]^{\top}$ and initializes zero residual banks $\Delta^c\in\mathbb{R}^{|\mathcal{C}_{\mathrm{eval}}|\times d}$, $\Delta^a\in\mathbb{R}^{M\times d}$, and $\Delta^o\in\mathbb{R}^{N\times d}$. Full PRISM reconstructs every migrated prototype, whereas Adaptive PRISM-Lite reconstructs only $c\in\mathcal{C}_i$. For $c=(a,o)$, HiKAM computes
\begin{equation}
    r_c=\Delta^c_c+\lambda_a\Delta^a_a+\lambda_o\Delta^o_o,\,
    \tilde{t}_c=\operatorname{Norm}(t_c+w_c r_c).
    \label{eq:hier_residual}
\end{equation}
Attribute and object residuals are shared across related compositions, while $\Delta^c_c$ preserves pair-specific adaptation. The separate banks provide separability, and the adaptive update weight (AUW) scales their joint correction.

For test batch $\mathcal{B}$, the image-conditioned AUW is
\begin{equation}
    w_{i,c}=\frac{1}{1+\exp(\theta\,\bar{v}_i^{\top}\bar{t}_c)},
    \,
    w_c=\frac{1}{|\mathcal{B}|}\sum_{i\in\mathcal{B}}w_{i,c}.
    \label{eq:update_weight}
\end{equation}
AUW gives larger corrections to poorly aligned prototypes while keeping aligned prototypes near the base representation. The composition logit is $\ell^{\mathrm{comp}}_{i,c}=\gamma\bar{v}_i^{\top}\tilde{t}_c$.

The shared primitive banks also update attribute and object prototypes $u_a$ and $q_o$:
\begin{equation}
    \tilde{u}_a=\operatorname{Norm}(u_a+\eta_a\Delta^a_a),
    \,
    \tilde{q}_o=\operatorname{Norm}(q_o+\eta_o\Delta^o_o),
    \label{eq:primitive_proto}
\end{equation}
Here, Eq.~\eqref{eq:update_weight} gives $\eta_a$ and $\eta_o$ using primitive prototypes and temperature $\theta_p$. The primitive logits are $\ell^a_{i,a}=\gamma\bar v_i^{\top}\tilde u_a$ and $\ell^o_{i,o}=\gamma\bar v_i^{\top}\tilde q_o$. For $c=(a,o)$, the branch fuses factorized and primitive-composed evidence:
\begin{equation}
    \small
    \begin{aligned}
    \ell^{\mathrm{PI}}_{i,(a,o)}
    =\frac{\gamma}{2}\bar{v}_i^{\top}\Big[
    \tilde{u}_a+\tilde{q}_o
    +\operatorname{Norm}
    (\tilde{t}_{(a,o)}+\lambda_a\tilde{u}_a+\lambda_o\tilde{q}_o)
    \Big].
    \end{aligned}
    \label{eq:primitive_evidence}
\end{equation}
Primitive confidence and composition agreement gate noisy evidence. Let $H_i^a$ and $H_i^o$ be the attribute and object entropies normalized by $\log M$ and $\log N$, and set $\rho_i=1-\frac{1}{2}(H^a_i+H^o_i)$. For $\hat c_i=\arg\max_c\ell^{\mathrm{comp}}_{i,c}$, $A_i$ is the mean probability assigned to $\hat c_i$ by the factorized and primitive-composed logits in Eq.~\eqref{eq:primitive_evidence}. The text-side logit is
\begin{equation}
    \small
    \ell^{\mathrm{text}}_{i,c}
    =\ell^{\mathrm{comp}}_{i,c}
    +\mu\,\operatorname{clip}
    \left(\rho_i^{\kappa}\left[(1-\omega)+\omega A_i\right],g_{\min},1\right)
    \ell^{\mathrm{PI}}_{i,c},
    \label{eq:text_logit}
\end{equation}
where $\mu$ is the primitive-logit weight, $\kappa$ the confidence power, $\omega$ the agreement weight, and $g_{\min}$ the minimum gate.

\subsection{Priority Visual Memory Queue and PQI}
\label{sec:priority_visual_memory_queue}

PVMQ supplies bounded test-stream visual evidence, while VisualKAM adapts its class-indexed prototypes before cache matching. For each composition $c$, PRISM maintains $\mathcal{Q}_c=\{(f_j,h_j)\}_{j=1}^{K_c}$, where $f_j$ is a normalized feature and $h_j$ its prediction entropy. Each feature enters its predicted top-1 queue; overflow removes the highest-entropy entry. This classwise policy caps memory, favors confident evidence, and prevents cross-class eviction. The memory prototypes are
\begin{equation}
    \begin{aligned}
    m_c
    &=\operatorname{Norm}\left(
    \frac{1}{|\mathcal{Q}_c|}
    \sum_{(f,h)\in\mathcal{Q}_c} f
    \right),\\
    \tilde{m}_j
    &=\operatorname{Norm}(m_j+\bar{w}_j\Delta^v_{y_j}),
    \end{aligned}
    \label{eq:visual_kam}
\end{equation}
where $\Delta^v$ is the visual residual bank, $y_j$ the prototype label, and $\bar w_j$ the batch-averaged AUW (Eq.~\eqref{eq:update_weight}) for $m_j$. For nonempty prototypes $\mathcal{M}=\{(\tilde{m}_j,y_j)\}_{j=1}^{J}$, the cache logit is
\begin{equation}
    \ell^{\mathrm{cache}}_{i,c}
    =\alpha\sum_{j:y_j=c}
    \exp\left[-\beta(1-\bar{v}_i^{\top}\tilde{m}_j)\right],
    \label{eq:cache_logit}
\end{equation}
where $\alpha$ controls cache strength and $\beta$ controls matching sharpness. The final prediction logit is $\ell_{i,c}=\ell^{\mathrm{text}}_{i,c}+\ell^{\mathrm{cache}}_{i,c}$.

\paragraph{Prototype Queue Initialization (PQI).}
PQI initializes PVMQ before reliable pseudo-labels by retaining low-entropy seen features and transferring their prototypes to unseen compositions through text similarity. With seen and unseen text prototypes $T_s,T_u$, seen visual prototypes $V_s$, and transfer temperature $\tau_m$,
\begin{equation}
    \small
    P_{s\rightarrow u}=\operatorname{Softmax}_{s}\left(\frac{T_sT_u^{\top}}{\tau_m}\right),
    \,
    V_u=\operatorname{Norm}(P_{s\rightarrow u}^{\top}V_s).
    \label{eq:virtual_unseen}
\end{equation}
\subsection{Online Objective and Inference}
\label{sec:online_objective}

\begin{table*}[!t]
    \centering
    \caption{Closed-world task-level comparison of complete CZSL systems on UT-Zappos, MIT-States, and C-GQA. The best and second-best AUC/HM values are bold and underlined, respectively.}
    \label{tab:closed_world_result_table}
    \resizebox{\textwidth}{!}{
    \begin{tabular}{l Accc Accc Accc}
        \toprule
        \multirow{2}{*}{Methods}
        & \multicolumn{4}{c}{UT-Zappos}
        & \multicolumn{4}{c}{MIT-States}
        & \multicolumn{4}{c}{C-GQA} \\
        \cmidrule(lr){2-5} \cmidrule(lr){6-9} \cmidrule(lr){10-13}
        & AUC & HM & Seen & Unseen
        & AUC & HM & Seen & Unseen
        & AUC & HM & Seen & Unseen \\
        \midrule
        DFSP~\cite{DFSP} \conf{CVPR'23}
        & 36.0 & 47.2 & 66.7 & 71.7
        & 20.6 & 37.3 & 46.9 & 52.0
        & 10.5 & 27.1 & 38.2 & 32.0 \\
        Troika~\cite{Troika} \conf{CVPR'24}
        & 41.7 & 54.6 & 66.8 & 73.8
        & 22.1 & 39.3 & 49.0 & 53.0
        & 12.4 & 29.4 & 41.0 & 35.7 \\
        PLID~\cite{PLID} \conf{ECCV'24}
        & 38.7 & 52.4 & 67.3 & 68.8
        & 22.1 & 39.0 & 49.7 & 52.4
        & 11.0 & 27.9 & 38.8 & 33.0 \\
        IMAX~\cite{IMAX} \conf{TPAMI'25}
        & 40.6 & 54.2 & 69.3 & 70.7
        & 21.9 & 39.1 & 48.7 & 53.8
        & 12.8 & 29.8 & 39.7 & 35.8 \\
        ClusPro~\cite{ClusPro} \conf{ICLR'25}
        & 46.6 & 58.5 & 70.7 & 76.0
        & \textbf{23.8} & \textbf{40.7} & 52.1 & 54.0
        & 14.9 & 32.8 & 44.3 & 37.8 \\
        TOMCAT~\cite{TOMCAT} \conf{NeurIPS'25}
        & \underline{48.3} &\underline{60.2} &74.5 & 72.8
        & 22.6 & 39.5 & 50.3 & 53.0
        & 16.0 & 34.0 & 45.3 & 40.1 \\
        FlowComposer~\cite{He2026FlowComposer} \conf{CVPR'26}
        & 46.8 & 58.6& 71.1& 74.9
        & \underline{23.5} & \underline{40.2} & 51.5 & 53.2
        & 15.9 & 34.0 & 44.8 & 40.7 \\
        I2CD~\cite{yuan2026i2cd} \conf{AAAI'26}
        & 46.4 & 57.9 & 69.7 & 77.1
        & 23.3 & 40.0 & 50.6 & 54.1
        & \underline{16.3} & \underline{35.0} & 44.1 & 41.2 \\
        \midrule
        \textbf{Full PRISM (Ours)}
        & \textbf{53.7} & \textbf{64.8} & 79.0 & 73.8
        & 23.3 & \underline{40.2} & 50.7 & 53.7
        & \textbf{16.7} & \textbf{35.4} & 45.4 & 41.1 \\
        \bottomrule
    \end{tabular}
    }

    \vspace{4pt}
    \caption{Open-world task-level comparison of complete CZSL systems on UT-Zappos, MIT-States, and C-GQA.}
    \label{tab:open_world_result_table}
    \resizebox{\textwidth}{!}{
    \begin{tabular}{l Accc Accc Accc}
        \toprule
        \multirow{2}{*}{Methods}
        & \multicolumn{4}{c}{UT-Zappos}
        & \multicolumn{4}{c}{MIT-States}
        & \multicolumn{4}{c}{C-GQA} \\
        \cmidrule(lr){2-5} \cmidrule(lr){6-9} \cmidrule(lr){10-13}
        & AUC & HM & Seen & Unseen
        & AUC & HM & Seen & Unseen
        & AUC & HM & Seen & Unseen \\
        \midrule
        DFSP~\cite{DFSP} \conf{CVPR'23}
        & 30.3 & 44.0 & 66.8 & 60.0
        & 6.8 & 19.3 & 47.5 & 18.5
        & 2.4 & 10.4 & 38.3 & 7.2 \\
        Troika~\cite{Troika} \conf{CVPR'24}
        & 33.0 & 47.8 & 66.4 & 61.2
        & 7.2 & 20.1 & 48.8 & 18.7
        & 2.7 & 10.9 & 40.8 & 7.9 \\
        PLID~\cite{PLID} \conf{ECCV'24}
        & 30.8 & 46.6 & 67.6 & 55.5
        & 7.3 & 20.4 & 49.1 & 18.7
        & 2.5 & 10.6 & 39.1 & 7.5 \\
        IMAX~\cite{IMAX} \conf{TPAMI'25}
        & 32.3 & 47.5 & 68.4 & 57.3
        & 7.6 & 21.4 & 50.2 & 18.6
        & 2.6 & 11.2 & 38.7 & 7.9 \\
        ClusPro~\cite{ClusPro} \conf{ICLR'25}
        & 39.5 & 54.1 & 71.0 & 66.2
        & \underline{9.3} & \underline{23.0} & 51.2 & 22.1
        & 3.0 & 11.6 & 41.6 & 8.3 \\
        TOMCAT~\cite{TOMCAT} \conf{NeurIPS'25}
        & \underline{43.7} & \underline{57.9} & 74.1 & 65.8
        & 8.2 & 21.7 & 49.2 & 21.0
        & 4.2 & 14.2 & 45.1 & 10.6 \\
        FlowComposer~\cite{He2026FlowComposer} \conf{CVPR'26}
        & 35.5 & 51.0 & 70.1 & 61.2
        & 7.5 & 20.3 & 50.4 & 19.0
        & 3.5 & 12.6 & 43.5 & 10.2\\
        I2CD~\cite{yuan2026i2cd} \conf{AAAI'26}
        & 35.6 & 50.2 & 69.1 & 62.1
        & 8.2 & 21.4 & 50.6 & 20.6
        & \underline{4.3} & \underline{14.9} & 44.0 & 10.9 \\
        \midrule
        \textbf{Full PRISM (Ours)}
        & \textbf{49.7} & \textbf{62.4} & 78.2 & 69.1
        & \textbf{11.0} & \textbf{26.4} & 50.4 & 26.4
        & \textbf{5.5} & \textbf{17.6} & 44.7 & 14.0 \\
        \bottomrule
    \end{tabular}
    }
\end{table*}
PRISM optimizes only test-time residual variables. Given final logits $\ell_{i,c}$,
\begin{equation}
    \begin{aligned}
    \mathcal{L}_{\mathrm{ent}}
    &=\frac{1}{B}\sum_{i=1}^{B}
    \left[-\sum_{c\in\mathcal{C}_i}p_{i,c}\log p_{i,c}\right],
    \\
    p_{i,c}&=\operatorname{softmax}(\ell_i)_c.
    \end{aligned}
    \label{eq:entropy_loss}
\end{equation}
The primitive branch minimizes $\mathcal{L}_{\mathrm{prim}}=\frac{1}{2B}\sum_i[H(\operatorname{softmax}(\ell^a_i))+H(\operatorname{softmax}(\ell^o_i))]$. Visual and text prototypes align through
\begin{equation}
\small
    \mathcal{L}_{\mathrm{align}}
    =\frac{1}{2}
    \left[
    \operatorname{CE}(R,I)+\operatorname{CE}(R^{\top},I)
    \right],
    \,
    R_{j,k}=\gamma\,\tilde{m}_j^{\top}\tilde{t}_k,
    \label{eq:align_loss}
\end{equation}

where $I$ is the identity target over classes with nonempty memory. PRISM also regularizes primitive residuals with $\mathcal{L}_{\mathrm{orth}}=\frac{1}{MN}\sum_{m,n}(\operatorname{Norm}(\Delta^a_m)^{\top}\operatorname{Norm}(\Delta^o_n))^2$. The online objective is
\begin{equation}
    \mathcal{L}_{\mathrm{tta}}
    =\mathcal{L}_{\mathrm{ent}}
    +\lambda_{\mathrm{prim}}\mathcal{L}_{\mathrm{prim}}
    +\lambda_{\mathrm{align}}\mathcal{L}_{\mathrm{align}}
    +\lambda_{\mathrm{orth}}\mathcal{L}_{\mathrm{orth}}.
    \label{eq:tta_loss}
\end{equation}
At each step, the selected route defines $\mathcal{C}_i$; PRISM extracts features, migrates active prototypes, fuses primitive and cache logits, predicts, and updates residuals with Eq.~\eqref{eq:tta_loss}. Adaptive PRISM-Lite masks scores outside $\mathcal{C}_i$. CLIP remains frozen, while residuals and PVMQ persist across batches.

\section{Experiments}
\label{sec:experiments}
\subsection{Experimental Setup}
\label{sec:experimental_setup}

\textbf{Datasets.}
We evaluate PRISM on UT-Zappos, MIT-States, and C-GQA~\cite{ut-zappos,mit-states,C-GQA} using the standard compositional-split-natural protocol. Held-out Unseen compositions are disjoint from Seen training compositions, while their attribute and object primitives remain shared. Dataset statistics are provided in the Technical Supplement.

\textbf{Evaluation protocol and metrics.}
Closed-world evaluation uses the benchmark Seen and Unseen pairs, whereas open-world evaluation expands the candidates to $\mathcal{A}\times\mathcal{O}$ with benchmark feasibility calibration. Following standard generalized CZSL evaluation~\cite{AUC_purushwalkam2019taskdrivenmodularnetworkszeroshot,Co-CGE,TOMCAT}, we vary an Unseen-composition bias and report AUC, HM, Seen, and Unseen. Formal definitions are provided in the Technical Supplement.

\textbf{Online evaluation protocol.}
All online experiments use prequential batch-1 predict-then-update evaluation. Each image is scored before adaptation, the updated state is carried forward, and test labels are used only for evaluation. Matched No TTA, Text TTA only, and Full PRISM runs share the checkpoint, candidate space, and test-stream order.

\begin{table*}[!t]
    \centering
    \caption{Closed-world ablation study on UT-Zappos, MIT-States, and C-GQA.}
    \label{tab:tta_ablation_result_table}
    \tiny
    \renewcommand{\arraystretch}{0.8}
    \setlength{\tabcolsep}{1.8pt}
    \resizebox{\textwidth}{!}{
    \begin{tabular}{lcccccccccccccccc}
        \toprule
        \multirow{2}{*}{Variant}
        & \multirow{2}{*}{HiKAM}
        & \multirow{2}{*}{Prim.}
        & \multirow{2}{*}{PVMQ}
        & \multirow{2}{*}{PQI}
        & \multicolumn{4}{c}{UT-Zappos}
        & \multicolumn{4}{c}{MIT-States}
        & \multicolumn{4}{c}{C-GQA} \\
        \cmidrule(lr){6-9}
        \cmidrule(lr){10-13}
        \cmidrule(lr){14-17}
        & & & &
        & AUC & HM & Seen & Unseen
        & AUC & HM & Seen & Unseen
        & AUC & HM & Seen & Unseen \\
        \midrule
        
        w/o TTA
        &  &  &  &
        & 43.6 & 55.5 & 68.7 & 74.3
        & 22.2 & 39.3 & 49.1 & 52.9
        & 15.2 & 33.1 & 43.9 & 39.6 \\
        
        Text TTA only
        & \checkmark &  &  &
        & 38.4 & 52.6 & 66.8 & 66.4
        & 22.2 & 39.2 & 49.1 & 53.0
        & 15.2 & 33.1 & 44.7 & 38.7 \\
        
        w/o PVMQ
        & \checkmark & \checkmark &  &
        & 44.5 & 57.3 & 71.3 & 71.2
        & 22.2 & 39.1 & 49.0 & 53.0
        & 15.6 & 33.4 & 45.3 & 38.9 \\
        
        w/o PQI
        & \checkmark & \checkmark & \checkmark &
        & 48.9 & 60.5 & 75.4 & 72.6
        & 22.3 & 39.0 & 49.3 & 53.3
        & 15.6 & 33.4 & 45.2 & 39.2 \\
        
        \midrule
        
        \textbf{Full PRISM}
        & \checkmark & \checkmark & \checkmark & \checkmark
        & \textbf{53.7} & \textbf{64.8} & \textbf{79.0} & \textbf{73.8}
        & \textbf{23.3} & \textbf{40.2} & \textbf{50.7} & \textbf{53.7}
        & \textbf{16.7} & \textbf{35.4} & \textbf{45.4} & \textbf{41.1} \\
        
        \bottomrule
    \end{tabular}}
\end{table*}

\textbf{Implementation details.}
PRISM uses a frozen CLIP ViT-L/14 backbone and is implemented in PyTorch. Efficiency measurements use a single NVIDIA GeForce RTX 3090 GPU. Complete training, adaptation, calibration, and hardware settings are provided in the Technical Supplement.

\textbf{Baselines.}
Published CZSL results position complete systems, with TOMCAT~\cite{TOMCAT} as the closest online comparator. Because checkpoints, offline training, and adaptation protocols vary across papers, adaptation-specific evidence comes from matched No TTA and component ablations. Extended baseline results are provided in the Technical Supplement.

\subsection{Main Results}
\label{sec:main_results}

Table 1 and Table 2 report the published task-level comparison against complete CZSL systems under the closed- and open-world settings, respectively.

\textbf{Closed-world results.} PRISM attains 53.7/64.8 AUC/HM on UT-Zappos and 16.7/35.4 on C-GQA, exceeding the strongest listed results by 5.4/4.6 and 0.4/0.4 points, respectively. On MIT-States, PRISM reaches 23.3/40.2 AUC/HM, matching FlowComposer in HM and trailing ClusPro by 0.5 AUC points, a gap that likely reflects the dataset's substantial label noise, estimated at roughly 70\% by a human-rater study~\cite{atzmon2020causal}, rather than as direct evidence of a systematic limitation in the online adaptation path. These margins characterize complete systems with different offline training and adaptation protocols; Table 3 subsequently isolates PRISM's own adaptation-specific contribution.

\textbf{Open-world results.} PRISM records 49.7/62.4 AUC/HM on UT-Zappos and 11.0/26.4 on MIT-States, exceeding the strongest listed results by 6.0/4.5 and 1.7/3.4 points, respectively. On C-GQA, whose candidate space expands from 7,767 closed-world pairs to 278,362 Cartesian-product candidates, PRISM reaches 5.5/17.6 AUC/HM, compared with the strongest listed values of 4.3/14.9.

PRISM thus occupies the strongest task-level position among the complete systems listed in Tables 1 and 2, with its largest margins recorded in the more demanding open-world setting. The following matched ablation evaluates the contribution of its online path under controlled conditions.

\subsection{Ablation Study}
\label{sec:ablation_studies}
To evaluate the contribution of PRISM's core online-adaptation components, we conduct ablation studies examining HiKAM, the primitive branch, PVMQ, and PQI on UT-Zappos, MIT-States, and C-GQA in the closed-world setting, with all variants sharing the checkpoint and test stream. Table~\ref{tab:tta_ablation_result_table} presents the comparative results.
Relative to No TTA, Full PRISM improves AUC/HM by 10.1/9.3, 1.1/0.9, and 1.5/2.3 points on UT-Zappos, MIT-States, and C-GQA, respectively. Text-only TTA yields no consistent gain, confirming that text migration alone is insufficient. Removing PVMQ causes the largest visual-side degradation, reaching 9.2/7.5 points on UT-Zappos and 1.1/2.0 on C-GQA, while PQI provides a further 4.8/4.3-point gain on UT-Zappos. Complete component and regularization ablations are reported in the Technical Supplement. Overall, the ablation confirms that PRISM's gains arise from jointly adapting primitive-shared text prototypes and visually grounded memory evidence.

The Technical Supplement further reports a matched audit that separately measures insertion-time pseudo-label accuracy and final PVMQ purity. Default purity ranges from 24.43\% to 60.94\%; dual-confidence gating improves purity by 11--23 points but retains only 37--47\% of entries, with small dataset-dependent changes in AUC/HM. These results expose a reliability--coverage trade-off, indicating that memory utility also depends on compositional coverage, visual diversity, and primitive-level agreement.

\subsection{Efficiency Analysis}
\label{sec:efficiency_analysis}

\begin{table}[!t]
    \centering
    \caption{C-GQA closed- and open-world profiling; Adaptive candidate counts are calibrated/realized.}
    \label{tab:app_prism_lite_profile}
    \scriptsize
    \setlength{\tabcolsep}{2pt}
    \renewcommand{\arraystretch}{0.9}
    \resizebox{\columnwidth}{!}{
    \begin{tabular}{cccccccc}
    \toprule
    Setting & Mode & Candidates & Recall (\%) & Img/s & Mean (ms) & AUC & HM \\
    \midrule
    
    \multirow{4}{*}{\rotatebox[origin=c]{90}{Closed}}
    & Full PRISM    & 7,767 & 100.00 & 2.35 & 423.86 & 16.69 & 35.41 \\
    & Fixed Lite (100) & 100   & 90.92  & 15.32 & 64.48 & 15.67 & 34.16 \\
    & Fixed Lite (200) & 200   & 95.21  & 14.83 & 66.63 & 15.65 & 34.04 \\
    & \textbf{Adaptive PRISM-Lite} & 148/152.3 & 93.55 & 13.11 & 76.28 & 15.93 & 34.41 \\
    
    \midrule
    
    \multirow{4}{*}{\rotatebox[origin=c]{90}{Open}}
    & Full PRISM    & 278,362 & 100.00 & 1.84 & 543.89 & 5.45 & 17.58 \\
    & Fixed Lite (100) & 100     & 71.81  & 10.63 & 94.09 & 5.10 & 16.42 \\
    & Fixed Lite (200) & 200     & 78.01  & 10.57 & 94.63 & 5.13 & 16.74 \\
    & \textbf{Adaptive PRISM-Lite} & 1,734/1,797.1 & 93.33 & 8.95 & 111.74 & 5.24 & 17.14 \\
    
    \bottomrule
    \end{tabular}}
\end{table}
All routes share the checkpoint, modules, objective, online updates, and batch size; Table~\ref{tab:app_prism_lite_profile} reports recurring post-calibration/PQI cost in both settings. Adaptive PRISM-Lite averages 152.3/1,797.1 candidates and retains 93.55/93.33\% recall in closed/open worlds. Relative to Full PRISM, it delivers $5.58\times$/$4.87\times$ higher throughput and 82.0/79.5\% lower latency, with closed/open AUC/HM losses of 0.76/1.00 and 0.21/0.44 points. Fixed Top-100/200 are slightly faster, but recall falls from 90.92/95.21\% closed-world to 71.81/78.01\% open-world. Adaptive PRISM-Lite therefore achieves a strong accuracy--efficiency trade-off by recalibrating its candidate budget to the evaluation universe. These gains concern recurring candidate-dependent computation and leave one-time PQI unchanged. Coverage assumes validation--test rank exchangeability; calibration, timing, and rank-shift details appear in the Technical Supplement.

\subsection{Fine-Grained Long-Tail Evaluation}
\label{sec:finegrained_longtail_evaluation}
\begin{table}[!t]
    \centering
    \caption{Long-tail analysis on UT-Zappos under closed- and open-world protocols.}
    \label{tab:longtail_analysis}
    \setlength{\tabcolsep}{2pt}
    \resizebox{\columnwidth}{!}{
    \begin{tabular}{cccc|ccc|ccc}
        \toprule
        \multirow{2}{*}{{Analysis}}
        & \multirow{2}{*}{Group}
        & \multirow{2}{*}{\#Comp.}
        & \multirow{2}{*}{\#Images}
        & \multicolumn{3}{c|}{Closed-world}
        & \multicolumn{3}{c}{Open-world}\\
        \cline{5-10}
        &&&&
        \makecell[c]{No TTA} 
        & \makecell[c]{\textbf{PRISM}} 
        & \makecell[c]{$\mathbf{\Delta}$}
        &
        \makecell[c]{No TTA} 
        & \makecell[c]{\textbf{PRISM}} 
        & \makecell[c]{$\mathbf{\Delta}$}\\
        \midrule
        
        \multirow{4}{*}{$\mathbf{c}$}
        & All
        & 36 & 2914
        & 24.3 & \textbf{34.6} & \textbf{+10.3}
        & 24.1 & \textbf{33.1} & \textbf{+8.9}\\
        & Tail
        & 11 & 50
        & 13.6 & \textbf{32.1} & \textbf{+18.5}
        & 13.6 & \textbf{29.7} & \textbf{+16.1}\\
        & Body
        & 14 & 546
        & 19.2 & \textbf{20.9} & \textbf{+1.8}
        & 18.7 & \textbf{19.9} & \textbf{+1.2}\\
        & Head
        & 11 & 2318
        & 41.6 & \textbf{54.5} & \textbf{+12.9}
        & 41.5 & \textbf{53.2} & \textbf{+11.7}\\
        
        \midrule
        
        \multirow{3}{*}{$\mathbf{a}$}
        & Tail
        & 8 & 152
        & 26.1 & \textbf{37.1} & \textbf{+11.0}
        & 26.1 & \textbf{37.3} & \textbf{+11.2}\\
        & Body
        & 11 & 597
        & 14.9 & \textbf{30.0} & \textbf{+15.1}
        & 14.9 & \textbf{28.1} & \textbf{+13.1}\\
        & Head
        & 17 & 2165
        & 29.5 & \textbf{36.4} & \textbf{+6.9}
        & 29.2 & \textbf{34.3} & \textbf{+5.2}\\
        
        \midrule
        
        \multirow{3}{*}{$\mathbf{o}$}
        & Tail
        & 9 & 330
        & 19.9 & \textbf{27.0} & \textbf{+7.1}
        & 19.2 & \textbf{25.8} & \textbf{+6.6}\\
        & Body
        & 12 & 1429
        & 24.7 & \textbf{35.0} & \textbf{+10.3}
        & 24.6 & \textbf{34.0} & \textbf{+9.4}\\
        & Head
        & 15 & 1155
        & 26.7 & \textbf{38.8} & \textbf{+12.2}
        & 26.7 & \textbf{36.7} & \textbf{+10.0}\\
        
        \bottomrule
    \end{tabular}}
\end{table}
Fine-grained long-tail analysis separates adaptation gains on under-supported concepts from gains on frequent classes. We compare PRISM with No TTA on three datasets using class-averaged accuracy under closed- and open-world protocols, with both variants sharing the checkpoint, test stream, and grouping (composition Tail/Body/Head by test-image count; attribute/object groups by training frequency). Results on UT-Zappos are reported in Table~\ref{tab:longtail_analysis}.
In the closed-world setting, the largest gain occurs in composition Tail (+18.5, from 13.6\% to 32.1\%), exceeding the +10.3 overall gain, while Body and Head improve by +1.8 and +12.9. Attribute-tail and object-tail also rise by +11.0 and +7.1. The open-world protocol preserves this pattern: overall composition accuracy rises +8.9, Tail rises +16.1, and attribute-tail (+11.2) and object-tail (+6.6) remain positive, with most gains only 0.5-2.4 points below their closed-world counterparts.
These results identify correction of under-supported compositions and primitives as PRISM's main long-tail benefit: the Tail gain exceeds the overall gain in both worlds, showing that PRISM's advantage is concentrated on rare, under-supported cases rather than merely reinforcing well-supported Head classes.

\subsection{Cumulative Classification Accuracy Analysis}
\label{sec:cumulative_accuracy_analysis}

We track cumulative top-1 accuracy as each closed-world test stream progresses to examine whether online gains persist over time. Figure 3 shows that PRISM follows an overall upward trajectory and maintains a late-stream advantage over the frozen Base Model on all three datasets. These trajectories indicate that PRISM sustains its online gains without progressive performance degradation as the stream evolves.

\begin{figure}[t]
    \centering
    \setlength{\tabcolsep}{2pt}
    \begin{tabular}{ccc}
        \includegraphics[width=0.14\textwidth]{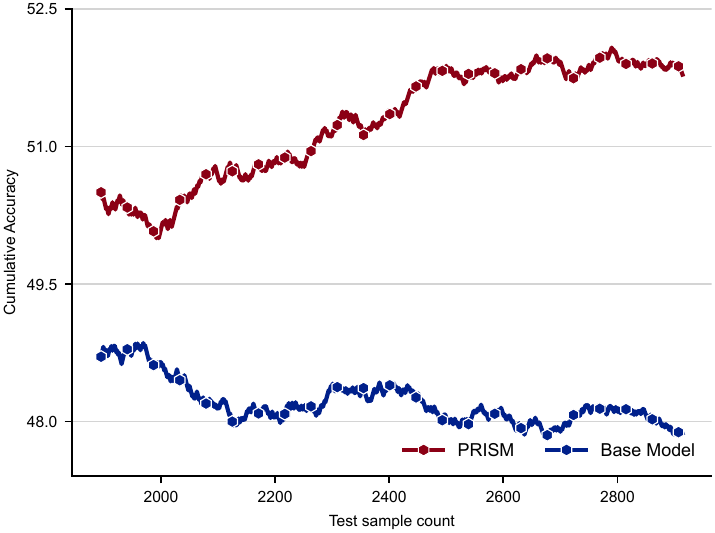} &
        \includegraphics[width=0.14\textwidth]{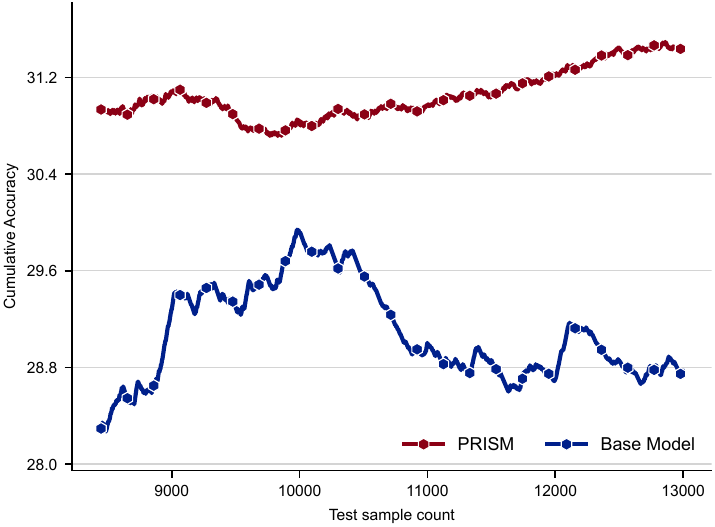} &
        \includegraphics[width=0.14\textwidth]{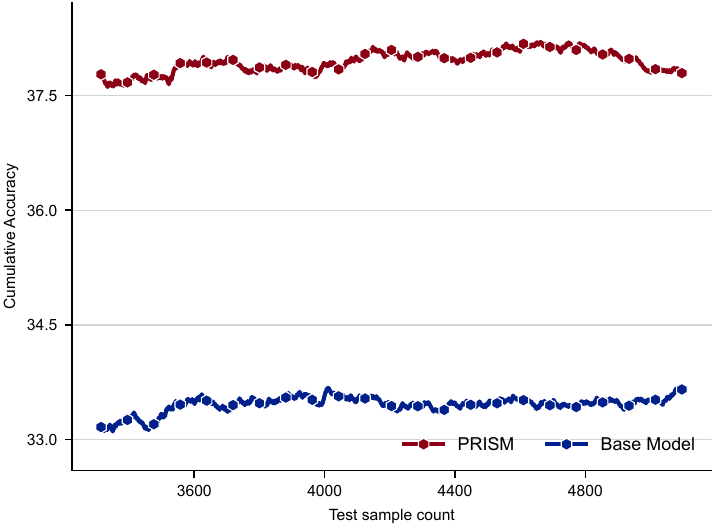} \\
        {\scriptsize UT-Zappos} &
        {\scriptsize MIT-States} &
        {\scriptsize C-GQA}
    \end{tabular}
    \caption{Closed-world cumulative top-1 accuracy with increasing test sample size on three datasets.}
    \label{fig:closed_cumulative_accuracy}
\end{figure}

\subsection{Non-Stationary Primitive-Block Evaluation}
\label{sec:nonstationary_stream_evaluation}

We conduct a non-stationary stream experiment using primitive-block ordering to impose prolonged object- or attribute-level skew. Full protocol details and quantitative results are reported in the Technical Supplement, where PRISM maintains stable end-task performance and consistent AUC/HM gains over No TTA across all three datasets. The Technical Supplement further analyzes shared-residual drift under primitive-block streams, providing a conditional theoretical basis for these empirical findings.

\subsection{Hyperparameter Robustness}
\begin{figure}[!t]
    \centering
    \setlength{\tabcolsep}{1pt}
    \begin{tabular}{ccc}
        \includegraphics[width=0.33\linewidth]{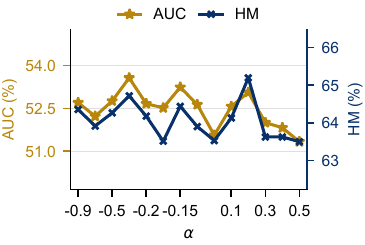} &
        \includegraphics[width=0.33\linewidth]{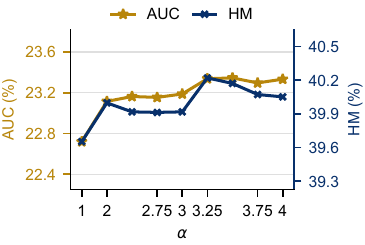} &
        \includegraphics[width=0.33\linewidth]{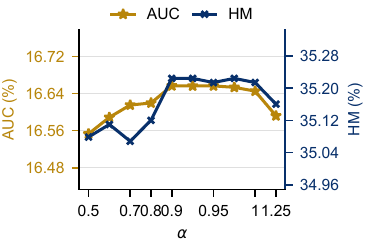} \\
        {\scriptsize (a) UT-Zappos} &
        {\scriptsize (b) MIT-States} &
        {\scriptsize (c) C-GQA}
    \end{tabular}
    \caption{Closed-world sensitivity analysis of $\alpha$.}
    \label{fig:alpha_tuning}
\end{figure}

\begin{figure}[!t]
    \centering
    \setlength{\tabcolsep}{1pt}
    \begin{tabular}{ccc}
        \includegraphics[width=0.33\linewidth]{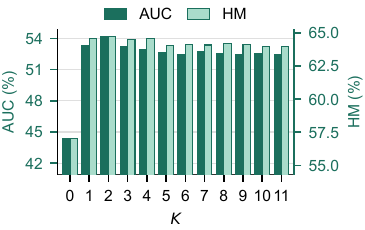} &
        \includegraphics[width=0.33\linewidth]{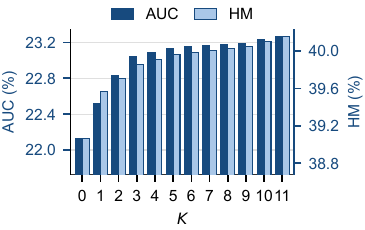} &
        \includegraphics[width=0.33\linewidth]{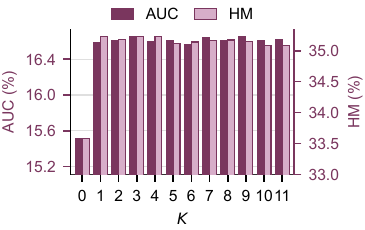} \\
        {\scriptsize (a) UT-Zappos} &
        {\scriptsize (b) MIT-States} &
        {\scriptsize (c) C-GQA}
    \end{tabular}
    \caption{Closed-world sensitivity analysis of Priority Visual Memory Queue capacity $K$.}
    \label{fig:k_tuning}
\end{figure}

We conducted a sensitivity analysis on the hyper-parameters cache strength $\alpha$ and queue capacity $K$ in the closed-world setting by varying one parameter at a time while keeping the others fixed at their benchmark-specific settings, with results shown in Figure~\ref{fig:alpha_tuning} and Figure~\ref{fig:k_tuning}. 
$\alpha$ shows relatively larger but smooth fluctuations as AUC/HM vary by at most 2.2/1.7 points on UT-Zappos, 0.6/0.6 on MIT-States, and 0.1/0.2 on C-GQA . In contrast, varying $K$ has little to no effect on AUC/HM within the evaluated range. The remaining hyperparameters, including cache sharpness $\beta$ and the primitive residual weights $\lambda_a$ and $\lambda_o$, exhibit similarly negligible impact, with detailed results provided in the Technical Supplement. These results collectively demonstrate that PRISM is robust to hyperparameter choices.

Although PRISM is relatively insensitive to $\alpha$, selecting an appropriate value can further reduce manual tuning. We therefore investigate whether $\alpha$ can be determined from unlabeled stream statistics. Across the three benchmarks, the selected $\alpha$ follows the ordering of the base model's mean prediction entropy $\bar{H}$ over the first few hundred unlabeled stream samples, summarized by the empirical mapping
\begin{equation}
    \alpha \approx \lambda \cdot \bigl(\exp(\tau \cdot (\bar{H} - H_0)) - 1\bigr),
    \label{eq:alpha_heuristic}
\end{equation}
where $H_0{=}1.75$, $\tau{=}4.5$, and $\lambda{=}0.45$ yield estimates within 0.06 of the selected values. As a label-free cold-start cue, this early-window estimate supplies a candidate fixed $\alpha$ before online adaptation begins.

\section{Conclusion}
\label{sec:conclusion}
PRISM extends primitive sharing in compositional zero-shot learning from offline representation learning to unlabeled test streams, jointly addressing open-world candidate expansion, long-tail evidence scarcity, and the recurring cost of online inference. Its hierarchical composition-, attribute-, and object-level residuals (HiKAM) propagate stream evidence across related concepts, while PVMQ and PQI transfer visual evidence to prototypes. Full PRISM applies this adaptive core to exhaustive inference and achieves the highest reported open-world AUC/HM across all three benchmarks, along with the strongest closed-world results on UT-Zappos and C-GQA. On C-GQA, it obtains 5.5/17.6 AUC/HM over 278,362 candidates, versus 4.3/14.9 for the strongest alternative, with matched same-checkpoint ablations confirming positive online-path contributions on all three closed-world benchmarks.
Aggregate scores understate these gains: the largest improvements occur on rare, under-supported concepts. On UT-Zappos, Tail composition accuracy rises by 18.5 points in the closed world and 16.1 points in the open world, exceeding overall gains of 10.3 and 8.9 points.
Adaptive PRISM-Lite preserves this online adaptation while reducing repeated computation in large candidate spaces. It delivers $4.87\times$ higher throughput and 79.5\% lower latency with 93.33\% candidate recall, incurring only 0.21/0.44-point AUC/HM losses relative to Full PRISM. These results establish primitive-shared online adaptation as an effective accuracy--efficiency framework for large-scale and long-tail CZSL. The current evidence remains bounded by validation--test exchangeability and the availability of reusable primitive support; extending calibrated coverage to shifted streams and improving visual-memory reliability are important directions for future work.

\bibliography{aaai2027}

\end{document}